\documentclass{elsart}
\usepackage{amssymb}
\usepackage{amsmath}
\usepackage{natbib}
\usepackage[french,english]{babel}
\usepackage[latin1]{inputenc}
\usepackage{graphicx}
\usepackage{url}
\usepackage{algorithmic}
\usepackage{algorithm}
\usepackage{supertabular}
\usepackage{fancyvrb}
\usepackage{rotating}

\newcommand{\vdata}{\mathbf{x}}
\newcommand{\vproto}{\mathbf{p}}
\newcommand{\cluster}{\mathcal{C}}
\newcommand{\affect}{f}
\newcommand{\ddataproto}{\gamma}
\newcommand{\protoset}{A}
\newcommand{\aff}{\mathrm{aff}}
\newcommand{\itermax}{L}
\newcommand{\diss}{d}
\newcommand{\partition}{\mathcal{P}}
\newcommand{\referent}{\mathcal{R}}
\newcommand{\referentset}{\mathcal{R}}

\makeatletter
\def\@keywordheading{{\it Mots clés : }}
\def\emails@name{Adresses électroniques }
\def\ps@copyright{\let\@mkboth\@gobbletwo
  \def\@oddhead{}%
  \let\@evenhead\@oddhead
  \def\@oddfoot{}%
  \let\@evenfoot\@oddfoot
}
 \def\and{\unskip~et~}
\renewcommand{\ALG@name}{Algorithme}
\makeatother

\begin{document}
\selectlanguage{french}
\sloppy
\begin{frontmatter}

\title{Une adaptation des cartes auto-organisatrices pour des données décrites
  par un tableau de dissimilarités}

\author[axis]{A{\"\i}cha El Golli\corauthref{cor}}, \ead{Aicha.ElGolli@inria.fr}
\author[axis]{Fabrice Rossi}, %
\ead{Fabrice.Rossi@inria.fr}
\author[lita,axis]{Brieuc Conan-Guez}, %
\ead{Brieuc.Conan-Guez@iut.univ-metz.fr} \and
\author[axis]{Yves Lechevallier} %
\ead{Yves.Lechevallier@inria.fr}%

\address[axis]{Projet AxIS, INRIA, Domaine de Voluceau, Rocquencourt,
  B.P. 105,\\ 78153 Le Chesnay Cedex, France}
\address[lita]{LITA EA3097, Université de Metz, Ile du Saulcy, F-57045 Metz,
France}

\corauth[cor]{Auteur à contacter.}

\begin{abstract}
  De nombreuses méthodes d'analyse des données ne sont applicables qu'aux
  données qui peuvent être représentées par un nombre fixé de valeurs
  numériques, alors que la plupart des observations issues de problèmes réels
  ne se trouvent pas naturellement sous cette forme. Il est alors
  indispensable d'adapter les méthodes prévues pour le cas vectoriel à des
  données plus complexes. Une solution très souple et très générale consiste à
  construire une mesure de (dis)similarité, basée sur le savoir des experts du
  domaine concerné, qui permette de comparer deux à deux les données
  étudiées. On construit alors des méthodes d'analyse qui ne travaillent qu'à
  partir du tableau de dissimilarités résumant les données.

  Dans cet article, nous proposons une adaptation des cartes
  auto-organisatrices de Kohonen à tout type de données pour lesquelles une
  mesure de (dis)similarité est définie. L'algorithme proposé est une
  adaptation de la version \emph{batch} de la méthode employée pour les
  données classiques.

  Nous validons notre méthode sur une application réelle, l'analyse de l'usage
  du site web de l'Institut National de Recherche en Informatique et
  Automatique à partir de fichiers log de ses serveurs.
\end{abstract}
\begin{keyword}
  Classification non supervisée \sep Projection non linéaire \sep Cartes
  auto-organisatrices \sep Dissimilarités \sep Web Usage Mining
\end{keyword}
\end{frontmatter}
\footnotetext{Article à paraître dans la Revue de Statistique Appliquée}

\pagebreak[3]

\selectlanguage{english}
\hrule height 0.4pt
\vskip 8pt \nopagebreak[4]
\textbf{Abstract}

Many data analysis methods cannot be applied to data that are not
represented by a fixed number of real values, whereas most of real
world observations are not readily available in such a format.
Vector based data analysis methods have therefore to be adapted in
order to be used with non standard complex data. A flexible and
general solution for this adaptation is to use a (dis)similarity
measure. Indeed, thanks to expert knowledge on the studied data,
it is generally possible to define a measure that can be used to
make pairwise comparison between observations. General data
analysis methods are then obtained by adapting existing methods to
(dis)similarity matrices.

In this article, we propose an adaptation of Kohonen's Self
Organizing Map (SOM) to (dis)similarity data. The proposed
algorithm is an adapted version of the vector based batch SOM.

The method is validated on real world data: we provide an analysis of the
usage patterns of the web site of the Institut National de Recherche en
Informatique et Automatique, constructed thanks to web log mining method.

\textit{Key words: } Clustering, Nonlinear projection, Self Organizing Map,
Dissimilarity, Web Usage Mining

\vskip 10pt
\hrule height 0.4pt

\selectlanguage{french}

\section{Introduction}
Dans de nombreuses applications, les observations ne sont pas naturellement
représentées sous forme d'un nombre fixé de valeurs numériques, i.e., sous
forme de vecteurs de $\Rset^p$. Les données réelles peuvent en effet être de
taille variable, être décrites par des variables qui ne sont pas directement
comparables, ne pas être numériques, etc. On peut évoquer par exemple les
données textuelles, les données semi-structurées (e.g., les documents XML),
les données fonctionnelles \citep{Ram97} et les données symboliques
\citetext{intervalles, distributions, etc., cf \citealt{Sod99}}. Or, beaucoup
de méthodes d'analyse de données ont été construites en exploitant les
propriétés de $\Rset^p$, plus particulièrement les opérations classiques qu'on
peut appliquer aux vecteurs de cet espace : combinaisons linéaires, produit
scalaire et norme. Pour être appliquées à des données non vectorielles, les
méthodes en question doivent être modifiées et adaptées. Dans certaines
conditions, notamment pour les données symboliques, il est possible de mettre
en {\oe}uvre des techniques de représentations numériques \citep{Aur02,Aur03,
  Chav02, Chav03, Ver00} pour se ramener au cas vectoriel. De la même façon,
l'utilisation d'un opérateur de projection permet d'associer à des
données fonctionnelles une représentation vectorielle
satisfaisante \citep{Ram97}.

Ces approches ont cependant une portée limitée, en particulier car la
représentation numérique induit souvent une perte d'information. Une
alternative particulièrement fructueuse consiste à s'appuyer sur la définition
de mesures de (dis)similarités entre données complexes et à généraliser les
méthodes classiques d'analyse de données au cas de tableaux de
(dis)similarités. L'avantage évident de cette stratégie est de séparer la
construction d'algorithmes d'analyse du choix de la représentation des
données. Cela permet de proposer une implémentation unique d'un algorithme
d'analyse qui pourra être utilisée avec toute sorte de données, à condition de
pouvoir calculer une (dis)similarités entre les observations. L'algorithme et
son implémentation deviennent alors universels. De plus, il devient possible
de faire appel à des experts pour définir la (dis)similarité utilisée, sans
que ceux-ci n'aient besoin de connaître l'algorithme utilisé : seule la
pertinence de la comparaison entre les données est importante pour les
experts. On peut ainsi réutiliser les nombreux travaux réalisés dans le
domaine, comme par exemple \citet[chapitre 8 \emph{``Similarity and
  dissimilarity''}, pages 139-197]{Sod99} pour les données symboliques et
\citet{Sha99} pour les données semi-structurées.

Nous proposons dans le présent article une adaptation de
l'algorithme des cartes auto-organisatrices \citep{Koh95} au cas
de données décrites par un tableau de dissimilarités. Les cartes
auto-organisatrices sont un instrument très utile pour l'analyse
exploratoire des données car elles combinent une classification
avec une projection non linéaire. Le principe fondamental de cette
méthode est de représenter un ensemble d'observations grâce à des
prototypes (aussi appelés référents) organisés selon une structure
fixée \emph{a priori}. Les prototypes doivent réaliser une bonne
quantification des données d'origine : chaque observation est
affectée à un prototype, ce qui définit une partition des données.
Chaque prototype est alors représentatif de la classe des
observations qui lui sont affectées, ce qui donne aux cartes
auto-organisatrices des aspects similaires à l'algorithme des
\emph{k-means} \citep{Mac67}. De plus, chaque prototype tient compte des
observations affectées aux classes voisines. Enfin, la structure
\emph{a priori} permet une représentation graphique des
prototypes, en général sur un plan. Cette représentation
s'apparente à une projection non linéaire, car la cohérence des
classes et la représentativité des prototypes autorisent à
considérer ces derniers comme les projetés des données d'origine.
Les cartes auto-organisatrices fournissent ainsi une alternative à
des méthodes classiques de projections, linéaires comme les
méthodes factorielles \citep{Hot33} ou non linéaires comme le
\emph{Multi Dimensional Scaling} \citep{Tor52} ou Isomap
\citep{IsomapScience2000}.

La généralisation que nous proposons s'adapte à toutes données pour lesquelles
une dissimilarité peut être définie. Elle est donc plus générale que les
extensions spécifiques des cartes auto-organisatrices qui ont été proposées
pour certaines données complexes comme les données symboliques \citep{Boc01},
les données structurées de type séries temporelles, arbres ou graphes
\citep{HammerEtAl2004StructuredNeurocomputing}, les chaînes de caractères
\citep{Somervuo2004OnlineSymbol}, les données qualitatives \citep{CottrellEtAl2004Qualitative,CottrellLetremy2005Survey} et les
données fonctionnelles \citep{RossiConanGuezElGolliESANN2004SOMFunc}.

Dans la section \ref{sectionSOM}, nous commencerons par rappeler l'algorithme
\emph{batch} des cartes auto-organisatrices. Nous montrerons dans la section
\ref{sectionDSOM} comment l'algorithme peut être adapté au cas de données
décrites uniquement par un tableau de dissimilarités. Nous conclurons cet
article par une application de notre algorithme à un problème réel d'analyse
de l'usage d'un site Web, présenté dans la section \ref{sectionWUM}. Nous
verrons à cette occasion que l'algorithme proposé donne des résultats très
satisfaisants en terme d'analyse du site web de l'Institut National de
Recherche en Informatique et Automatique (INRIA).

\section{Les cartes auto-organisatrices}\label{sectionSOM}
\subsection{Principe général}
L'algorithme des cartes auto-organisatrices de Kohonen \citep{Koh95}, abrégé
en SOM pour \emph{Self Organizing Map}, est à la fois un algorithme de
projection non linéaire et un algorithme de classification. Il associe à des
données d'origine (appartenant en général à un espace de grande dimension) un
ensemble de prototypes organisés selon une structure de faible dimension (en
général deux) choisie \emph{a priori}. Chaque prototype représente un
sous-ensemble des données d'origine qu'on peut considérer comme une classe.
L'organisation des prototypes, et donc des classes, est imposée par la
structure \emph{a priori}, mais elle est aussi contrainte par les données
elles-mêmes de sorte que la représentation graphique des prototypes réalise
une projection non linéaire des données qui préserve leur topologie.

Plus formellement, la structure (c'est-à-dire la \emph{carte}) est décrite par
un graphe $(C,\Gamma)$. $C$ désigne les $M$ \emph{neurones} de la carte.
Chaque neurone est associé à un prototype (aussi appelé référent du neurone)
et à une classe (on aura donc $M$ classes). L'organisation \emph{a priori}
provient de l'ensemble d'arêtes $\Gamma$ : deux neurones $c$ et $r$ sont
connectés directement et donc voisins dans la carte si $(c,r)\in\Gamma$. Cette
structure de graphe induit une distance discrète $\delta$ sur la carte : pour
tout couple de neurones $(c, r)$ de la carte, la distance $\delta(c, r)$ est
définie comme étant la longueur du plus court chemin entre $c$ et $r$.

Le but de l'algorithme SOM est, partant d'un ensemble de $N$
observations, les $\vdata_1,\ldots,\vdata_N$ (qui forment
l'ensemble $\Omega$), d'associer à chaque neurone $c\in C$ un
prototype $\vproto_c$ et un sous-ensemble $\cluster_c$ de
$\Omega$. On demande que les $(\cluster_c)_{c\in C}$ forment une
partition de $\Omega$ et que pour tout $c$, $\vproto_c$ représente
de façon satisfaisante les éléments de $\cluster_c$ (il s'agit
d'une mesure de qualité de la partition) : ceci correspond à
l'aspect classificatoire de l'algorithme SOM. De plus il faut que
la structure \emph{a priori} soit respectée, c'est-à-dire que si
$c$ et $r$ sont des neurones proches (au sens de la distance
$\delta$ induite par le graphe $\Gamma$), alors $\vproto_c$ doit représenter
correctement les éléments de $\cluster_r$ (de même pour $\vproto_r$ par
rapport à $\cluster_c$).

\subsection{Les cartes auto-organisatrices pour les données classiques} \label{SOMClassique}
Dans la section précédente, nous sommes volontairement restés très
vagues dans la description des données, des prototypes et de la
notion de proximité (ou de représentation satisfaisante), afin de
rappeler le principe général des cartes auto-organisatrices. Nous
allons maintenant rappeler en détail la version dite \emph{batch}
de l'algorithme pour des données classiques, c'est-à-dire des
éléments d'un espace vectoriel $\Rset^p$. Dans cette situation,
$\Omega$ est donc un ensemble de $N$ vecteurs de $\Rset^p$. Les
prototypes (les $\vproto_c$) sont aussi choisis dans $\Rset^p$.
Enfin, les données sont comparées au sens de la distance
euclidienne.

Pour formaliser la notion de respect de la structure \emph{a priori} et de
qualité de la partition, on utilise une fonction noyau $K$ de $\Rset^+$ dans
$\Rset^+$, décroissante et telle que $K(0)=1$ et $\lim_{x\rightarrow
  \infty}K(x)=0$ (en pratique, on utilise souvent $K(x)=e^{-x^2}$). Cette
fonction engendre une famille de fonctions, les $K^T$, définies
par $K^T(x)=K(\frac{x}{T})$. Le paramètre $T$ est analogue à une
température \citep{THI97,THI02} : quand $T$ est élevé, $K^T(x)$
reste proche de $1$ même pour de grandes valeurs de $x$ ; au
contraire une valeur faible engendre une fonction $K^T$ qui
décroît très vite vers 0. Le rôle de $K^T$ est de transformer la
distance discrète $\delta$ induite par la structure de graphe en
une fonction de voisinage plus régulière et paramétrée par $T$ :
on utilisera ainsi $K^T(\delta(c,r))$ comme mesure de proximité
effective entre les neurones $c$ et $r$. Pendant le déroulement de
l'algorithme SOM, la valeur de $T$ décroît afin d'assurer la
stabilisation de la solution obtenue.

La qualité de la partition $(\cluster_c)_{c\in C}$ et des prototypes associés,
les $(\vproto_c)_{c\in C}$, est alors donnée par l'énergie suivante
\citep{Cheng1997BatchSomConvergence}, qui doit 
être la plus faible possible :
\begin{equation}\label{eqEnergy}
E^T\left((\cluster_c)_{c\in C},(\vproto_c)_{c\in C}\right)=
\sum_{\vdata_i\in  \Omega} \sum_{c\in C}
K^T\left(\delta(\affect(\vdata_i),c)
\right)\left\|\vproto_c-\vdata_i\right\|^2,
\end{equation}
où $\affect$ désigne la fonction d'affectation, telle que
$\affect(\vdata_i) =c$ si $\vdata_i\in\cluster_c$. Pour simplifier la suite du
texte, on note $\partition=(\cluster_c)_{c\in C}$ la partition et
$\referent=(\vproto_c)_{c\in C}$ le système de prototypes associé.

Pour bien comprendre le sens de l'énergie, on peut la récrire de la façon
suivante :
\begin{equation}\label{eqEnergyBis}
E^T\left(\partition,\referent\right)=\sum_{r\in
  C}\sum_{\vdata_i\in\cluster_r}\sum_{c\in
  C}K^T\left(\delta(r,c)\right)\left\|\vproto_c-\vdata_i\right\|^2.
\end{equation}
Comme $K^T(0)=1$, on peut décomposer l'énergie en deux termes :
\begin{equation}\label{eqEnergyR}
E^T\left(\partition,\referent\right)_R=\sum_{r\in C}
\sum_{\vdata_i\in\cluster_r}\left\|\vproto_r-\vdata_i\right\|^2,
\end{equation}
et
\begin{equation}\label{eqEnergyS}
E^T\left(\partition,\referent\right)_S=\sum_{r\in C}
\sum_{c\neq r}\sum_{\vdata_i\in \cluster_c}
K^T\left(\delta(r,c)\right)\left\|\vproto_r-\vdata_i\right\|^2.
\end{equation}
Le terme $E^T\left(\partition,\referent\right)_R$
correspond à la distorsion utilisée dans les algorithmes de classification de
type \emph{k-means} \citep{Mac67}. Le terme $E^T\left(\partition,\referent\right)_S$ est lui spécifique aux cartes
auto-organisatrices. On voit qu'il impose au prototype du neurone $r$ de
représenter les observations qui ont été affectées à d'autres neurones. Un
défaut de représentation, c'est-à-dire une grande valeur pour
$\left\|\vproto_r-\vdata_i\right\|^2$, pèse d'autant plus lourdement dans
l'énergie que le neurone $r$ est proche (dans la structure \emph{a priori}) du
neurone $\affect(\vdata_i)$.

La minimisation de l'énergie $E^T\left(\partition,\referent\right)$ est un
problème d'optimisation combinatoire. On se contente en pratique d'une
solution sous-optimale obtenue par une heuristique. Une telle heuristique est
donnée par la version \emph{batch} de l'algorithme SOM \citep{Koh95} et par
ses variantes \citep{Heskes93,Cheng1997BatchSomConvergence,THI97}.  Les
variantes de l'algorithme alternent deux étapes distinctes, une étape
d'affectation (calcul de $\affect$) et une étape de représentation (calcul des
$\vproto_c$), ce qui le classe dans les algorithmes de type nuées dynamiques
\citep{did71,disi76,dig77}.

L'algorithme \emph{batch} initial \citep{Koh95} utilise une étape
d'affectation de type \emph{winner takes all} définie par
\begin{equation}
  \label{eqWinnerTakesAll}
\affect(\vdata)=\arg\min_{r\in C}\left\|\vproto_r-\vdata_i\right\|^2.  
\end{equation}
La convergence de cette version de l'algorithme a été étudiée dans
\citep{FortEtAl2001BatchSOM}, mais les résultats obtenus sont difficilement
extensibles aux cas des données décrites par des dissimilarités. 

Pour simplifier l'analyse, nous utilisons dans le présent article la variante
proposée par \citep{Heskes93}. L'étape d'affectation consiste ici à minimiser
$E^T\left(\partition,\referent\right)$ en considérant les prototypes fixés.
Comme dans l'algorithme \emph{batch} standard, l'étape de représentation
minimise la même énergie mais en considérant les classes comme fixées. Bien
que les deux optimisations soient réalisées de façon exacte, on ne peut pas
garantir que l'énergie est globalement minimisée par cet algorithme. Par
contre, si on fixe la structure de voisinage (pour $T$ fixé), l'algorithme
converge en un nombre fini d'étape vers un état stable
\citep{Cheng1997BatchSomConvergence}.

Comme l'énergie est une somme de termes indépendants, on peut remplacer les
deux problèmes d'optimisation par un ensemble de problèmes simples
équivalents.  La formulation de l'équation \ref{eqEnergy} montre que l'énergie
est construite comme la somme sur l'ensemble des observations d'une mesure
d'adéquation de $\Rset^{p}\times C$ dans $\Rset^+$ définie par :
\begin{equation}
  \label{eqDataProto}
\ddataproto^T(\vdata,r)=\sum_{c\in C}
K^T\left(\delta(r,c)\right)\left\|\vproto_c-\vdata\right\|^2,
\end{equation}
ce qui donne
\begin{equation}
  \label{eqEnergyDissDataProto}
E^T\left(\partition,\referent\right)=\sum_{\vdata_i\in\Omega}\ddataproto^T(\vdata_i,\affect(\vdata_i)).
\end{equation}
Pour optimiser $E^T$ en gardant les prototypes fixés, il suffit donc de
minimiser chacune des sommes indépendamment, ce qui amène à définir $\affect$
comme suit :
\begin{equation}
  \label{eqAffectOptimal}
\affect(\vdata)=\arg\min_{r\in C}\ddataproto^T(\vdata,r).
\end{equation}
De même, quand les classes sont fixées, l'optimisation de $E^T$ par rapport
aux prototypes s'obtient en minimisant l'énergie associée à chaque neurone,
à savoir :
\begin{equation}
  \label{eqEnergyNeuron}
E^T_c(\vproto)=\sum_{\vdata_i\in  \Omega}
K^T\left(\delta(\affect(\vdata_i),c) \right)\left\|\vproto-\vdata_i\right\|^2,
\end{equation}
ce qui revient à poser :
\begin{equation}
  \label{eqPrototypeOpti}
  \vproto_c=\arg\min_{\vproto\in\Rset^p}E^T_c(\vproto).
\end{equation}
Ce problème d'optimisation admet une solution simple définie comme la moyenne
pondérée des observations :
\begin{equation}
  \label{eqPrototype}
  \vproto_c=\frac{\sum_{\vdata_i\in  \Omega}
K^T\left(\delta(\affect(\vdata_i),c) \right)\vdata_i}{\sum_{\vdata_i\in
  \Omega} K^T\left(\delta(\affect(\vdata_i),c) \right)}
\end{equation}
La version \emph{batch} de l'algorithme SOM étudiée est alors celle décrite
dans l'algorithme \ref{algoSOM}.

\begin{algorithm}[htpb]
\caption{La version \emph{batch} de l'algorithme SOM}\label{algoSOM}
  \begin{algorithmic}[1]
  \STATE Choisir une valeur initiale pour les prototypes $(\vproto_c)_{c\in
    C}$ \COMMENT{Étape d'initialisation}
  \FOR{$l=1$ à $\itermax$}
    \FORALL[Étape d'affectation]{élément $\vdata$ de $\Omega$}
    \STATE calculer
\[
\affect(\vdata)=\arg\min_{r\in C}\ddataproto^T(\vdata,r)
\]
 \ENDFOR
  \FORALL[Étape de représentation]{neurone $c \in C$}
   \STATE calculer
\[
\vproto_c=\frac{\sum_{\vdata_i\in  \Omega}
K^T\left(\delta(\affect(\vdata_i),c) \right)\vdata_i}{\sum_{\vdata_i\in
  \Omega} K^T\left(\delta(\affect(\vdata_i),c) \right)}
\]
  \ENDFOR
\ENDFOR
  \end{algorithmic}
\end{algorithm}

Bien que cela n'apparaisse pas explicitement dans l'algorithme, la
température $T$ évolue en fonction de $l$, en général selon une
décroissance exponentielle avec $l$. D'autre part, nous ne
développons pas ici les différentes méthodes d'initialisation
disponibles pour le choix des valeurs initiales des prototypes :
de nombreuses variantes existent \citep[cf][]{Koh95,THI97,THI02}.

Enfin, il existe aussi de nombreuses variantes de l'algorithme des cartes
auto-organisatrices : nous nous focalisons sur la version \emph{batch} que
nous venons de décrire car elle s'adapte simplement au cas d'un tableau de
dissimilarités, comme nous allons le voir dans la section suivante. Notons que
le caractère déterministe des algorithmes \emph{batch} est central dans les
démonstrations de leurs convergences
\citep{Cheng1997BatchSomConvergence,FortEtAl2001BatchSOM}. L'algorithme
stochastique classique est beaucoup plus délicat à analyser \citep[cf par
exemple][]{CottrellEtAl1998SomTheory}.

\section{Adaptation à un tableau de dissimilarités}\label{sectionDSOM}
\subsection{Une carte et son énergie}
Comme nous l'avons indiqué en introduction, notre but est d'adapter les cartes
auto-organisatrices au cas de données décrites uniquement par l'intermédiaire
d'un tableau de dissimilarités. La différence fondamentale avec les
sections précédentes est que l'ensemble des observations $\Omega$ n'est plus
une partie de $\Rset^p$ mais un ensemble quelconque associé à une fonction
$\diss$, de $\Omega\times\Omega$ dans $\Rset^+$ qui vérifie les propriétés
suivantes :
\begin{itemize}
    \item $\diss$ est symétrique, i.e., $\diss(\vdata_i,\vdata_j) =
      \diss(\vdata_j,\vdata_i)$ ;
    \item $\diss$ est positive, i.e., $\diss(\vdata_i,\vdata_j)\geq 0$ ;
    \item $\diss(\vdata_i,\vdata_i)= 0$.
\end{itemize}
La fonction $\diss$ est donc une dissimilarité : $\diss(\vdata_i,\vdata_j)$
est d'autant plus faible que $\vdata_i$ et $\vdata_j$ sont ``semblables''.

Nous ne faisons aucune hypothèse structurelle sur $\Omega$, ce qui signifie
qu'aucune opération n'est possible sur cet ensemble, excepté le calcul de
$\diss$. Malgré cela, l'étude de l'équation \ref{eqEnergy} montre que la
notion de carte auto-organisatrice et d'énergie associée est généralisable à
la situation qui nous intéresse. En effet, l'énergie est définie à partir de
la distance euclidienne (au carré) entre les observations et les
prototypes. Le reste de l'équation \ref{eqEnergy} ne fait apparaître que la
structure \emph{a priori} et la partition des données. Il est donc tentant de
remplacer la distance dans l'équation \ref{eqEnergy} par la dissimilarité
définie sur $\Omega$. Ceci n'est possible que si les prototypes sont
contraints à être des éléments de $\Omega$.

Dans certaines situations, cette dernière contrainte peut être
considérée comme trop forte car $\Omega$ peut être un échantillon
peu représentatif de l'espace de départ. Nous proposons donc de
généraliser la notion de prototype ou de référent d'un neurone :
au lieu d'associer au neurone $c$ un unique prototype $\vproto_c$
(choisi donc dans $\Omega$), nous lui associons un sous-ensemble
$\protoset_c$ contenant $q$ éléments distincts de $\Omega$. Le
paramètre $q$ est déterminé en fonction du problème : si la valeur
$q=1$ conduit à des résultats difficiles à interpréter, on peut
augmenter $q$ afin de réduire les effets de l'échantillonnage.

L'énergie d'une carte auto-organisatrice ainsi définie est donc :
\begin{equation}\label{eqEnergyDiss}
E^T\left((\cluster_c)_{c\in C},(\protoset_c)_{c\in C}\right)=
\sum_{\vdata_i\in  \Omega} \sum_{c\in C}
K^T\left(\delta(\affect(\vdata_i),c)
\right)\sum_{\vdata_j\in\protoset_c}\diss(\vdata_i,\vdata_j).
\end{equation}
Comme dans le cas classique, on utilisera dans la suite du texte les notations
$\partition=(\cluster_c)_{c\in C}$ et $\referent=(\protoset_c)_{c\in C}$.

\subsection{Interprétation de l'énergie}\label{interenergie}
Cette énergie généralise clairement l'énergie de l'équation \ref{eqEnergy} au
cas d'un tableau de dissimilarités. On peut cependant s'interroger sur sa
signification. Comme pour le cas classique, elle se décompose en une partie
qui mesure la qualité de la classification et une autre qui impose le respect
de la structure \emph{a priori}.

La qualité de la classification est donc définie par :
\begin{equation}
  \label{eqEnergyDissClassif}
 E^T\left(\partition,\referentset\right)_R=\sum_{r\in C}
\sum_{\vdata_i\in\cluster_r}\sum_{\vdata_j\in\protoset_r}\diss(\vdata_i,\vdata_j),
\end{equation}
ce qui se réduit à
\begin{equation}
  \label{eqEnergyDissClassifOne}
 E^T\left(\partition,\referentset\right)_R=\sum_{r\in C}
\sum_{\vdata_i\in\cluster_r}\diss(\vdata_i,\vproto_r),
\end{equation}
quand $q=1$, c'est-à-dire le cas le plus simple. Le principe de cette mesure
est donc de dire qu'une partition est de bonne qualité si on peut trouver pour
chaque classe un prototype (ou un ensemble de prototypes) tels que les
éléments de la classe soient semblables au(x) prototype(s) (au sens de la
dissimilarité). De la même façon, le respect de la structure impose que le(s)
prototype(s) associé(s) à un neurone soi(en)t semblable(s) aux observations
associées aux neurones voisins.

Si la dissimilarité correspond à la distance euclidienne au carré,
l'énergie retenue est très satisfaisante. En effet, les propriétés
de la distance euclidienne (en particulier le théorème de
Huygens) font que les classes obtenues en minimisant l'énergie
sont compactes et bien séparées : les observations dans une même
classe sont proches les unes des autres (compacité) et sont
éloignées des observations des autres classes (séparation).

Dans le cas d'une dissimilarité métrique, c'est-à-dire qui vérifie l'inégalité
triangulaire (i.e., $\diss(u,w)\leq \diss(u,v)+\diss(v,w)$), la situation est
aussi satisfaisante. En effet, deux éléments d'une même classe sont proches
car $\diss(\vdata_i,\vdata_j)\leq
\diss(\vdata_i,\vproto_c)+\diss(\vproto_c,\vdata_j)$ et que par construction
les prototypes sont proches des observations de leur classe.

Par contre, pour une dissimilarité quelconque, les classes peuvent très bien
ne pas être compactes. Si l'inégalité triangulaire n'est pas vérifiée, les
observations affectées à un neurone peuvent être très proches du ou des
prototypes associés au neurone sans être proches entre elles.

En pratique cependant, le but des cartes auto-organisatrices est bien atteint
par la minimisation de l'énergie choisie. L'obtention de classes compactes
n'est pas en effet le but principal : il s'agit en fait de représenter
simplement (en deux dimensions en général) un ensemble de prototypes qui
peuvent être considérés comme représentatifs des observations
d'origine. L'énergie impose ici que les prototypes soient proches des
observations, ce qui correspond bien au but fixé.

\subsection{L'algorithme}
L'algorithme SOM adapté à un tableau de dissimilarités se définit alors sur le
modèle de la version \emph{batch} pour données classiques. On cherche en effet
à minimiser l'énergie (\ref{eqEnergyDiss}) de façon heuristique, en alternant
une étape d'optimisation par rapport à $\partition=(\cluster_c)_{c\in C}$
(l'affectation) avec une étape d'optimisation par rapport à
$\referentset=(\protoset_c)_{c\in C}$ (la représentation). Comme dans la
section précédente, ces optimisations se décomposent en des ensembles
d'optimisations simples.

On commence par transposer l'équation (\ref{eqDataProto}) au cas
des dissimilarités, ce qui donne :
\begin{equation}
  \label{eqDataProtoSet}
  \ddataproto^T(\vdata,r)=\sum_{c\in C}
K^T\left(\delta(r,c)\right)\sum_{\vdata_j\in\protoset_c}\diss(\vdata,\vdata_j),
\end{equation}
et
\begin{equation}
  \label{eqEnergyDissProto}
  E^T\left(\partition,\referentset\right)=\sum_{\vdata_i\in  \Omega} \ddataproto^T(\vdata_i,\affect(\vdata_i)).
\end{equation}
Exactement comme dans le cas classique, la phase d'affectation consiste donc à
trouver $r=\affect(\vdata_i)$ qui minimise $\ddataproto^T(\vdata_i,r)$.

Pour la phase de représentation, on définit l'énergie associée au
neurone $c$, sur le modèle de l'équation (\ref{eqEnergyNeuron}) :
\begin{equation}
  \label{eqEnergyNeuronDiss}
 E^T_c(\protoset)=\sum_{\vdata_i\in  \Omega}
K^T\left(\delta(\affect(\vdata_i),c)
\right)\sum_{\vdata_j\in\protoset}\diss(\vdata_i,\vdata_j).
\end{equation}
Optimiser $E^T$ par rapport à $(\protoset_c)_{c\in C}$ revient en fait à
optimiser les $E^T_c$ pour $c\in C$. $E^T_c$ associe à une partie quelconque
de $\Omega$ une énergie. Pour trouver le minimum de $E^T_c$ sur l'ensemble des
parties à $q$ éléments distincts, il suffit de trouver les $q$ éléments de
$\Omega$, qui donnent les $q$ plus petites valeurs pour
$E^T_c(\{\vdata\})$. Ceci peut se faire par force brute, c'est-à-dire en
calculant $E^T_c(\{\vdata\})$ pour tout $\vdata\in \Omega$.

\begin{algorithm}[htpb]
\caption{Les cartes auto-organisatrices pour tableau de dissimilarités}\label{algoDSOM}
  \begin{algorithmic}[1]
  \STATE Choisir une valeur initiale pour les prototypes $(\protoset_c)_{c\in
    C}$ \COMMENT{Étape d'initialisation}
  \FOR{$l=1$ à $\itermax$}
    \FORALL[Étape d'affectation]{élément $\vdata$ de $\Omega$}
    \STATE calculer
\[
\affect(\vdata)=\arg\min_{r\in C}\ddataproto^T(\vdata,r)
\]
 \ENDFOR
  \FORALL[Étape de représentation]{neurone $c \in C$}
   \STATE calculer
\[
\protoset_c=\arg\min_{\protoset\subset\Omega,\ |\protoset|=q}\sum_{\vdata_i\in  \Omega}
K^T\left(\delta(\affect(\vdata_i),c)
\right)\sum_{\vdata_j\in\protoset}\diss(\vdata_i,\vdata_j),
\]
où $|\protoset|$ désigne le cardinal de l'ensemble $\protoset$.
  \ENDFOR
\ENDFOR
  \end{algorithmic}
\end{algorithm}

En combinant les étapes présentées au dessus, on obtient
l'algorithme \ref{algoDSOM}. Quelques détails doivent être
précisés :
\begin{itemize}
\item l'initialisation est réalisée par un choix aléatoire des prototypes,
  c'est-à-dire des sous-ensembles $\protoset_c$ pour $c\in C$, qui sont
  choisis de sorte à être disjoints deux à deux ;
\item lors de l'optimisation de la phase d'affectation, il est possible que
  deux neurones ou plus réalisent la même énergie minimale, c'est-à-dire qu'on
  obtienne $c$ et $r$, tels que $c\neq r$ et
  $\ddataproto^T(\vdata,c)=\ddataproto^T(\vdata,r)$. Pour lever l'ambiguïté,
  on choisi pour $\affect(\vdata)$ la plus petite valeur pour $c$ qui réalise
  le minimum de $\ddataproto^T(\vdata,c)$ ;
\item de la même façon, il est possible que plusieurs choix pour le
  prototype $\protoset$ conduisent à la même énergie minimale pour
  $E^T_c(\protoset)$. On lève l'ambiguïté en conservant les observations
  d'indice le plus faible et en exigeant que des prototypes associés à des
  neurones distincts soient disjoints, i.e., $c\neq r$ implique
  $\protoset_c\cap\protoset_r=\emptyset$.
\end{itemize}
Notons pour finir que la preuve de convergence proposée dans
\citep{Cheng1997BatchSomConvergence} s'adapte parfaitement à notre algorithme,
même avec $T$ fixé. En effet, si on fixe la structure de voisinage (pour $T$
constant), chaque étape de l'algorithme réduit (au sens large) la valeur de
l'énergie $E^T$.  Comme celle-ci est toujours positive, elle converge. De
plus, la carte admet un nombre fini de configurations, puisque $\referentset$
est une liste de $M$ sous-ensembles disjoints de $\Omega$ (ensemble de
cardinal $N$) et qu'il existe un nombre fini de partition de $\Omega$ en $M$
classes. De ce fait, la valeur limite est atteinte en un nombre fini d'étapes.

En pratique cependant, le voisinage évolue et la convergence n'est pas assurée
théoriquement, même si l'expérience montre que l'algorithme se stabilise. De
plus, même en cas de structure de voisinage fixée, la configuration finale
n'est pas nécessairement celle qui minimise $E^T$ : l'algorithme ne peut que
diminuer la valeur initiale. Il est donc préférable de réaliser plusieurs
optimisations, en partant de configurations initiales aléatoires distinctes,
et de conserver la configuration finale d'énergie minimale. 

\subsection{Liens avec les travaux antérieurs}
Tout d'abord remarquons que si la fonction de voisinage $K^T$ est
de la forme $K^T(\vdata)=1$ si $\vdata=0$ et $K^T(\vdata)=0$
sinon, nous retrouvons la méthode de classification de type nuées
dynamique d'un tableau de proximité décrite dans \citep{celeu89}
(à condition de se limiter à une représentation par un seul
prototype, soit $q=1$). En effet, un tel choix pour $K^T$ revient
à ne tenir compte que de la partie de l'énergie définie par
l'équation (\ref{eqEnergyDissClassif}). La seconde partie
($E^T\left(\partition,\referentset\right)_S$) qui est dépendante
de la carte est ignorée. Nous retrouvons aussi, avec cette
méthode, les difficultés d'interprétation de l'énergie décrites
dans la section \ref{interenergie} bien que le critère optimisé
soit plus simple.

Deux adaptations des cartes auto-organisatrices aux tableaux de
dissimilarités ont été proposés. La première est assez proche de
la notre et est due à Kohonen et Somervuo
\citep{Koh96,KohonenSomervuo1998Symbol,KohonenSomervuo2002NonVectorial}.
La seconde est due à Graepel, Burger et Obermayer
\citep{GraepelEtAl1998GSOM,GraepelObermayer1999DSOM} et est plus
éloignée de notre solution.

L'algorithme de Kohonen et Somervuo \citep{KohonenSomervuo2002NonVectorial} est
assez proche de l'algorithme \ref{algoDSOM}. Les différences essentielles sont
les suivantes :
\begin{itemize}
\item Kohonen et Somervuo associent à chaque neurone un unique prototype ;
\item le critère d'affectation utilisé par Kohonen et Somervuo est simplement
  celui qui associe à une observation le neurone dont le prototype est le plus
  proche (au sens de la dissimilarité). Ceci pose quelques problèmes. En
  effet, certaines dissimilarités, comme la distance de Levenshtein
  \citep{Levenhstein} entre chaînes de caractères, sont à valeurs entières, ce
  qui favorise les égalités entre dissimilarités. Il est alors fréquent
  d'avoir le choix entre plusieurs neurones pour l'affectation d'une
  observation à un neurone. Réaliser un choix aléatoire entre les différentes
  possibilités introduit une source d'instabilité dans l'algorithme qui nuit à
  sa convergence. Somervuo et Kohonen proposent dans
  \citep{KohonenSomervuo1998Symbol,KohonenSomervuo2002NonVectorial} de résoudre
  le problème avec un algorithme d'affectation assez complexe qui consiste à
  calculer une forme de distance pondérée entre l'observation considérée et
  les prototypes des neurones voisins du neurone candidat. Le voisinage pris
  en compte grossit petit à petit tant qu'aucun neurone ne devient un
  vainqueur unique. En fait, l'algorithme proposé peut être considéré comme
  une version heuristique de notre critère d'affectation, au moins pour
  certaines familles de fonctions de voisinage $K^T$. Notons que notre critère
  est beaucoup moins sensible au problème de minima multiples ;
\item dans la phase de représentation, la recherche du nouveau prototype
  associé à un neurone $c$ se fait parmi l'ensemble des observations affectées
  au neurone $c$ (i.e., dans $\cluster_c$) et aux neurones voisins, alors que
  nous recherchons le(s) prototype(s) dans l'ensemble des observations. De
  plus, nous tenons compte des pondérations induites par la structure \emph{a
    priori} alors que Kohonen et Somervuo minimisent la somme des
  dissimilarités entre le prototype et les observations affectées au neurone
  courant et aux neurones voisins, sans pondération. On peut obtenir le même
  comportement dans notre algorithme en choisissant une famille de fonctions
  de voisinage $K^T$ un peu particulière.
\end{itemize}
Le principal défaut de l'algorithme de Kohonen et Somervuo réside dans le fait
qu'il ne correspond pas à une optimisation (même heuristique) d'un critère
donné. Bien qu'il permette d'obtenir des résultats satisfaisants, il est
difficile de savoir exactement ce qu'il fait et donc d'éviter de commettre des
erreurs d'interprétation, par exemple.

L'algorithme de Graepel, Burger et Obermayer est très différent du notre. En
revanche, il est aussi basé sur une heuristique d'optimisation d'un critère
d'énergie bien défini. Le critère retenu est construit à partir d'une mesure
de compacité des classes obtenues, en l'occurrence la somme de toutes les
dissimilarités entre les éléments d'une classe (normalisée pour éviter des
problèmes liés à des effectifs très différents dans des classes). L'avantage
de ce critère est qu'il garantit l'obtention de classes pertinentes, même si
la dissimilarité n'est pas métrique. En contrepartie, l'algorithme ne produit
pas de prototypes pour résumer une classe, ce qui limite les possibilités en
visualisation. De plus, il est très coûteux puisqu'il se comporte en $O(N^2M)$
(pour $N$ observations et $M$ neurones) contre $O(N^2+NM^2)$ pour le notre
\citep{RossiEtAl05FastDSOM}.

\subsection{Exemple de mise en {\oe}uvre pour des données simulées}
Pour valider le fonctionnement de l'algorithme, nous l'avons testé
avec la distance euclidienne (au carré) sur des données simulées.
On se donne par exemple un ensemble de 1000 observations disposées
uniformément sur un cylindre dans $\Rset^3$. La structure \emph{a
priori} est donnée par une grille bi-dimensionnelle de $21\times
3=63$ neurones.
\begin{figure}[htbp]
\begin{minipage}[b]{0.45\textwidth}
\centering
\includegraphics[height=5cm,angle=-90]{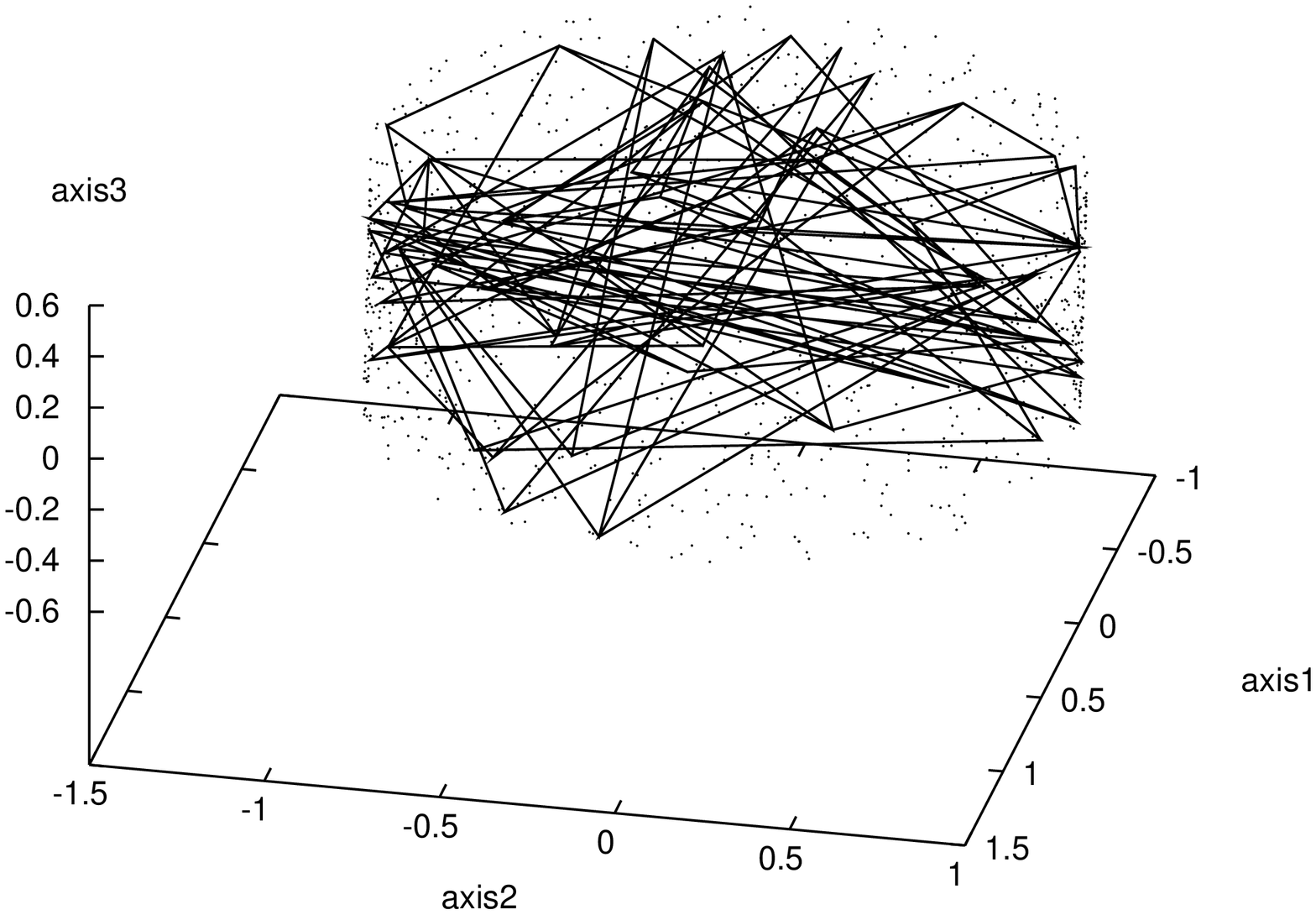}
\caption{La carte ($21 \times 3$ neurones) et le nuage des
points}\label{cylindre0}
\end{minipage}
\hfill
\begin{minipage}[b]{0.45\textwidth}
\centering
 \includegraphics[height=5cm,angle=-90]{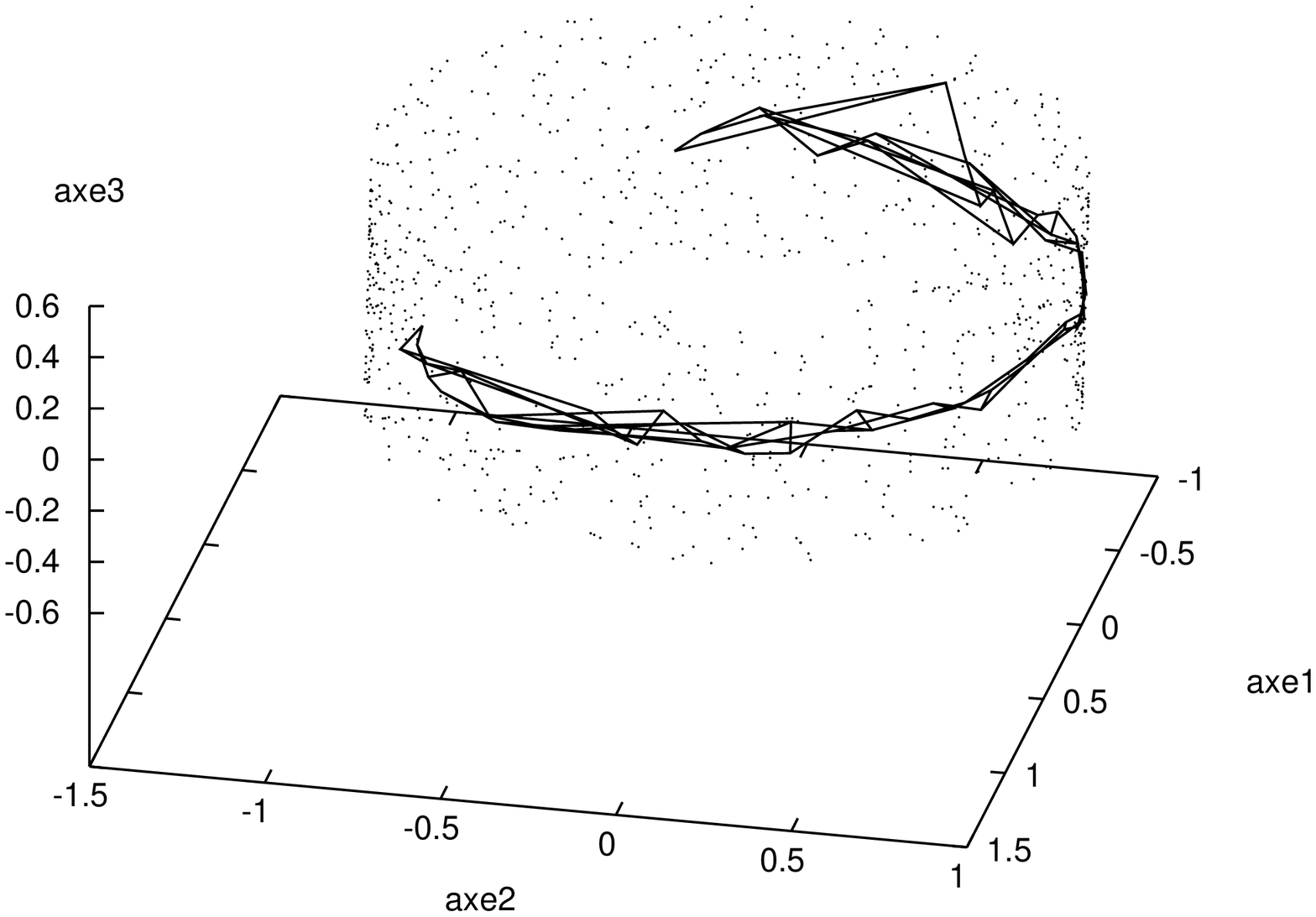}
 \caption{La carte après 50 itérations} \label{cylindre50}
 \end{minipage}
\end{figure}

\begin{figure}[htbp]
\begin{minipage}[b]{0.45\textwidth}
\centering
\includegraphics[height=5cm,angle=-90]{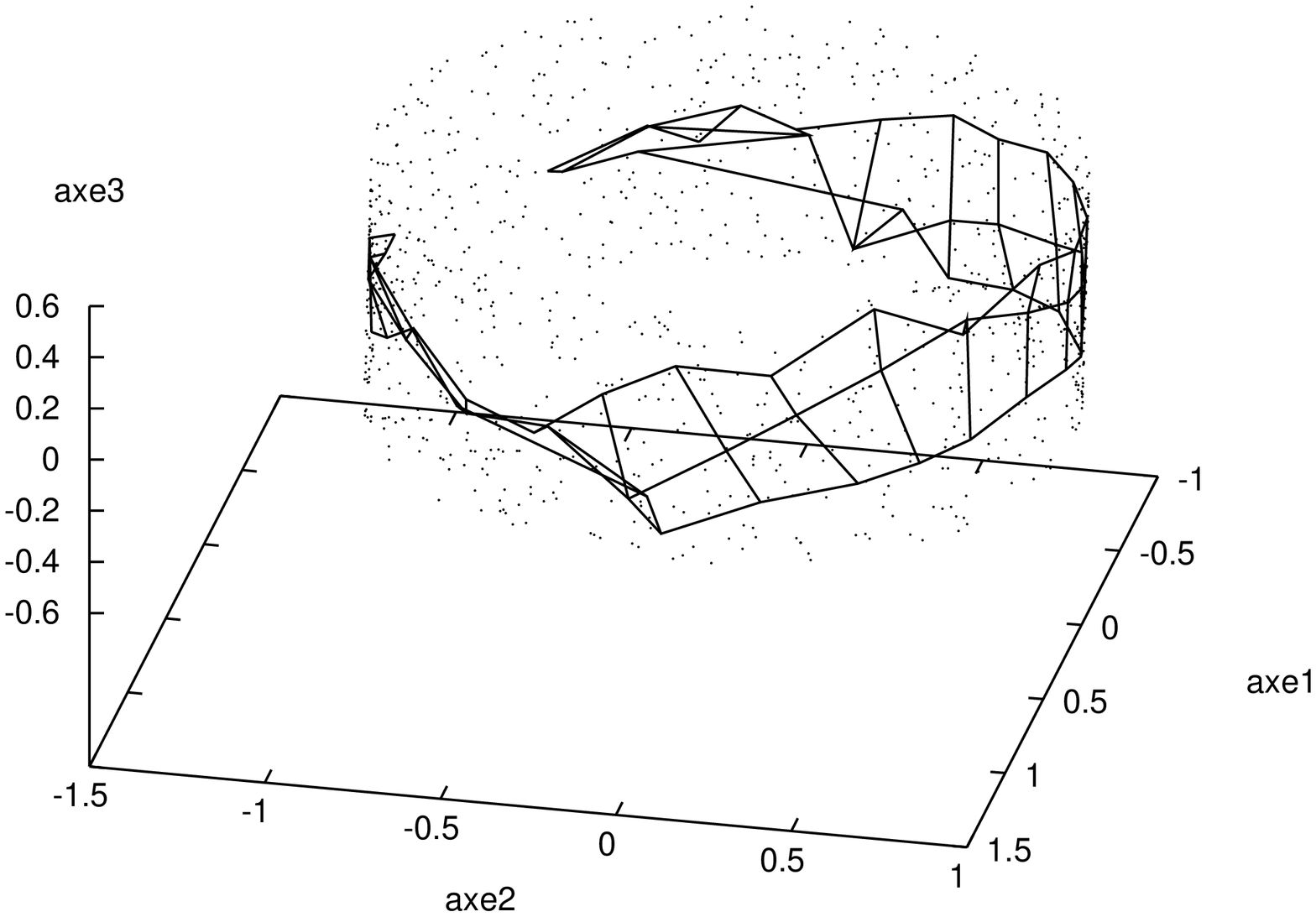}
\caption{La carte après 100 itérations} \label{cylindre100}
\end{minipage}
\hfill
\begin{minipage}[b]{0.45\textwidth}
\centering
\includegraphics[height=5cm,angle=-90]{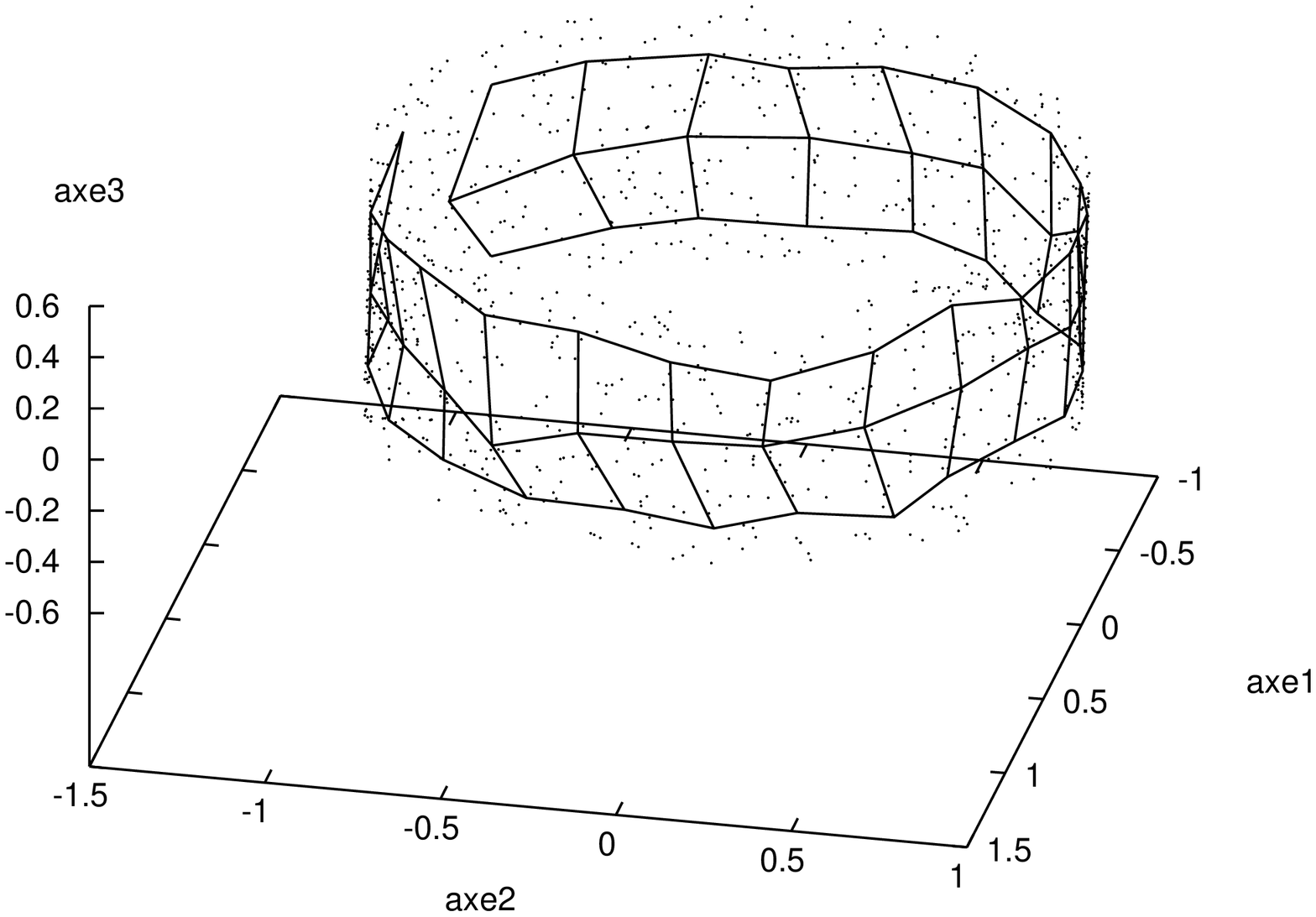}
\caption{La carte finale} \label{cylindrefinal}
\end{minipage}
\end{figure}

Dans les figures allant de \ref{cylindre0} à \ref{cylindrefinal}, nous
présentons les données et l'évolution de la carte durant l'apprentissage. La
figure \ref{cylindre0} présente la carte initiale sur le cylindre après
l'initialisation aléatoire des prototypes dans l'ensemble des observations (on
a choisi ici $q=1$). La carte finale présentée dans la figure
\ref{cylindrefinal} montre que la topologie des données à bien été retrouvée
par l'algorithme, que la carte est bien déployée et que la quantification est
tout à fait satisfaisante.

\section{Une application en analyse des usages d'un site Web}\label{sectionWUM}
\subsection{Introduction}
Dans cette section, nous proposons une illustration de l'intérêt de notre
méthode en l'utilisant pour analyser les usages d'un site Web. Les buts de
l'analyse des usages d'un site (le WUM, pour \emph{Web Usage Mining}) sont
nombreux \citetext{cf \citet{srivastava00web} pour une présentation
  synthétique des objectifs principaux du WUM}. Nous nous focaliserons ici sur
deux aspects, l'analyse des parcours d'un site par ses utilisateurs (les
navigations) qui vise à extraire des comportements typiques, et l'analyse de
la perception du site qui cherche à faire apparaître les similitudes entre les
différents contenus du site au sens de l'utilisation qu'en font les
internautes.

Avant de commencer à détailler les étapes de notre application, nous rappelons
les principaux concepts propres au WUM (les définitions sont inspirées de
celle du W3C et reprises de \citealt{Tan03})

Le contenu d'un site Web est un ensemble de documents (au sens large)
identifiés par des URLs (\emph{Uniform
Ressource Locators}, un cas particulier des \emph{Uniform Ressource
Identifiers}, \citealt[cf][]{URIRFC2396}). Une URL est de la forme simplifiée
suivante~: \url{http://<host>/<path>} (dans cet article, nous ne prendrons pas
en compte la partie recherche qui peut terminer une URL). La partie
\url{<host>} correspond au nom DNS du serveur considéré alors que la partie
\url{<path>} correspond au chemin d'accès au document demandé sur le serveur.
L'URL \url{http://www-sop.inria.fr/axis/} correspond ainsi au serveur
\url{www-sop.inria.fr} et au document \url{axis/} sur ce serveur. Nous ne
restreignons pas notre travail à l'analyse d'un site hébergé sur un seul
serveur, i.e. d'une partie \url{<host>} unique. Pour prendre en compte les
sites Web complexes utilisant plusieurs serveurs, nous considérons que
l'\url{<host>} peut varier.

La plupart des documents d'un site Web sont des pages au format (X)HTML
\citep{XHMTL1,HTML401} qui contiennent des hyperliens, c'est-à-dire des
références vers d'autres documents accessibles sur le Web (sous forme
d'URLs). Une page web regroupe parfois plusieurs documents, par exemple le
texte lui-même au format HTML, des images, des scripts externes, une applet
Java, etc.

Nous nous intéressons aux utilisateurs d'un site, c'est-à-dire à des personnes
qui consultent le site par l'intermédiaire d'un logiciel particulier appelé
navigateur. Le navigateur envoie des requêtes HTTP au serveur Web, en
désignant les URLs des documents que l'utilisateur souhaite consulter. On
appelle \textbf{session} l'ensemble des requêtes envoyées à un ou plusieurs
serveur(s) web par un utilisateur (par l'intermédiaire de son navigateur). Une
session est découpée en \textbf{navigations} (ou \textbf{visites}). Le
découpage est réalisé sur une base temporelle : quand l'écart temporel entre
deux requêtes d'une session dépasse un certain seuil (en général 30 minutes),
on considère qu'il y a rupture de la navigation. Une navigation est donc une
séquence de requêtes séparées d'au plus une certaine durée d'inaction.

\subsection{Fichiers log et pré-traitements}
Les données d'usage d'un site Web proviennent essentiellement des fichiers log
des serveurs concernés. Ceux-ci sont généralement écrits dans le format CLF
(\emph{Common Logfile Format}, \citealt{LuotonenCLF1995}) ou dans sa version
étendue qui comporte plus d'informations. Dans ce dernier format, chaque
requête vers le serveur Web est représentée par une ligne de la forme suivante
\citep{Gol00} :
\begin{center}
[Ip] [nom] [login] [date] [requête] [statut] [taille] [referrer]
[agent]
\end{center}
Les différents éléments de cette ligne sont les suivants :
\begin{center}
  \begin{supertabular}{p{2cm}p{11cm}}
    Ip& Adresse internet de provenance de la requête (en général, l'ordinateur
    de l'utilisateur). \\
    nom/login& L'accès à certaines ressources est contrôlé : le fichier log
    contient alors les identifiants nécessaires pour l'accès. \\
    date& Date et heure de réception de la requête.\\
    requête& La requête reçue par le serveur, dont nous allons essentiellement
    extraire l'URL du document demandé.\\
    statut& Code numérique précisant le statut de la requête au niveau du
    serveur (acceptée
    sans erreur, accès interdit, document inexistant, etc.)\\
    taille& Indique la taille du fichier retourné.\\
    referrer& URL du document dans lequel l'URL demandée a été trouvée
    (correspond à la structure hyper-textuelle des pages web) \\
    agent& Le navigateur et le type de système d'exploitation de l'utilisateur.\\
  \end{supertabular}
\end{center}
Voici un exemple de trace dans un fichier log :
\begin{center}
  \begin{Verbatim}[frame=single,fontsize=\small]
194.78.232.8 -- [10/Jan/2003:15:33:43 +0200] "Get /orion/liens.htm
HTTP/1.1" 200 1893 "http://www-sop.inria.fr/orion/index.html"
"Mozilla/4.0 (compatible; MSIE 5.0b1; Mac_PowerPC)
  \end{Verbatim}
\end{center}
On note les informations suivantes :
\begin{itemize}
\item 194.78.232.8 est l'adresse internet de l'utilisateur ;
\item la requête a été reçue le 10 Janvier 2003 à 15 heures, 33 minutes et 43
  secondes ;
\item l'URL demandée est \url{/orion/liens.htm} (il s'agit d'une URL relative
  à laquelle on doit ajouter le nom du serveur, ici \url{www-sop.inria.fr}) ;
\item la requête a été traitée sans erreur (c'est le statut 200) ;
\item le document renvoyé contenait 1893 octets ;
\item le lien vers ce document a été trouvé dans le document d'URL
  \url{http://www-sop.inria.fr/orion/index.html} ;
\item enfin, l'utilisateur travaillait sur un Macintosh (Mac\_PowerPC) avec le
  logiciel Internet Explorer (MSIE) en version 5.0b1.
\end{itemize}
Malgré leur apparente richesse, les fichiers log sont difficiles à exploiter
directement pour diverses raisons, dont voici une sélection :
\begin{itemize}
\item un serveur web reçoit en général des requêtes en provenance de plusieurs
  utilisateurs : les sessions sont donc entremêlées et il faut les
  reconstruire ;
\item les robots d'indexation des moteurs de recherche de type Google
  parcourent régulièrement la plupart des sites web, ce qui engendre de
  nombreuses requêtes automatiques : il faut supprimer ces requêtes pour se
  focaliser sur les requêtes engendrées par des humains ;
\item identifier la provenance d'une requête est difficile : certains
  utilisateurs passent par l'intermédiaire d'un \emph{proxy} qui réalise les
  requêtes à leur place. Des requêtes provenant d'utilisateurs distincts
  semblent alors venir d'une même adresse internet (et souvent d'un même
  agent, celui du \emph{proxy}). De plus, les ordinateurs sont souvent
  partagés entre utilisateurs dans certains contextes (université, \emph{cyber
    café}, etc.). Le problème inverse est aussi présent : il est fréquent que
  les utilisateurs non professionnels disposent seulement d'une adresse
  internet dynamique qui change donc à chaque connexion. Les requêtes d'un
  même utilisateur proviennent alors d'adresses internet distinctes ;
\item les sites recevant un trafic important utilisent en général plusieurs
  ordinateurs serveurs, et ont donc plusieurs fichiers log qu'il faut
  fusionner avant les traitements \citep{Tan03}.
\end{itemize}
Pour extraire du sens des logs, il est donc indispensable de procéder à une
étape de pré-traitement relativement complexe. Nous ne décrirons pas ici les
algorithmes retenus : nous avons utilisé les méthodes développées dans notre
équipe et décrites dans \citep{Doru04a,Doru04b}. Elles permettent d'extraire
des logs multi-serveurs les navigations réalisées par les utilisateurs, après
reconstruction de celles-ci et suppression des requêtes engendrées par les
robots. Le tableau \ref{tableData} donne un exemple des données obtenues grâce
à ces traitements.

\begin{table}[htbp]
\small
  \begin{center}
    \begin{sideways}
      \begin{tabular}{|c|c|c|c|c|l|}\hline
        Requête&Navigation&Session & Heure & Date & URL\\\hline\hline
        0 & 0 & 0 & \url{00:00:05} & \url{01/01/03} &  \url{http://www.inria.fr/}
        \\\hline
        1 & 0 & 0 & \url{00:00:25} & \url{01/01/03} & \url{http://www.inria.fr/inria/index.fr.html} \\\hline
        2 & 0 & 0 & \url{00:01:15} & \url{01/01/03} & \url{http://www.inria.fr/inria/unites.fr.html} \\\hline
        3 & 0 & 0 & \url{00:02:06} & \url{01/01/03} & \url{http://www.inria.fr/inria/liste-part.fr.html} \\\hline
        $\vdots$ & & & & & $\vdots$ \\\hline
        467 & 73 & 10 & \url{13:21:36} & \url{05/01/03} & \url{http://www.inria.fr/rapportsactivite/RA94/CROAP.3.4.2.html} \\\hline
        468 & 73 & 10 & \url{13:22:52} & \url{05/01/03} & \url{http://www.inria.fr/rapportsactivite/RA95/croap/node22.html} \\\hline
        469 & 73 & 10 & \url{13:25:01} & \url{05/01/03} & \url{http://www.inria.fr/rapportsactivite/RA96/croap/node22.html} \\\hline
        $\vdots$ & & & & & $\vdots$ \\\hline
      \end{tabular}
    \end{sideways}
\end{center}
  \caption{Tableau de données après pré-traitements}
  \label{tableData}
\end{table}

En fait, le pré-traitement peut être vu comme un enrichissement des fichiers
log par l'ajout des informations de navigation et de session. Ceci se traduit
dans l'exemple de la table \ref{tableData} par les colonnes portant ces noms
qui contiennent des identifiants uniques. L'ensemble des requêtes de la
session $k$ est constitué de l'ensemble des lignes dont la colonne session
vaut $k$ (de même pour le contenu d'une navigation). Bien entendu, le tableau
peut être enrichi par l'ajout d'autres informations disponibles dans les logs
comme le \emph{referrer}, l'\emph{agent}, etc. en fonction des besoins de
l'analyse.

Les analyses que nous allons décrire dans la suite de l'article sont toutes
basées sur un tableau de données de la forme qui vient d'être
présentée. Nous avons imposé les contraintes suivantes :
\begin{itemize}
\item nous n'analysons pas les requêtes vers des images ;
\item nous n'analysons que les requêtes correctes, c'est-à-dire avec un statut
  compris entre 200 et 399 ;
\item le seuil de rupture pour une navigation est de 30 minutes : une
  navigation est donc une suite de requêtes réalisées par un même utilisateur
  avec au plus 30 minutes entre deux requêtes ;
\item nous ne travaillons pas sur les sessions, mais uniquement sur les
  navigations : nous ne tenons donc pas compte du fait que plusieurs
  navigations peuvent provenir d'un même utilisateur. Ceci réduit les
  problèmes liés aux \emph{proxy}, aux adresses internet dynamiques et au
  partage d'ordinateurs.
\end{itemize}

\subsection{Prise en compte de la structure du site}\label{subsectionTopic}
Une des difficultés du WUM est qu'un site web de taille moyenne peut contenir
des milliers de documents. Même en analysant l'usage du site sur une longue
période, il reste difficile d'observer des comportements répétés sur lesquels
fonder une analyse : les navigations sont en général très différentes les
unes des autres, car chaque utilisateur se focalise sur la partie du site qui
l'intéresse. Pour pouvoir analyser le comportement des utilisateurs et leur
perception du site étudié, il est donc impératif de simplifier le
problème. Tout d'abord, nous supprimons l'aspect temporel des navigations (à
l'image de \citep{Mob02}, par exemple) : nous ne tiendrons donc pas compte dans
l'analyse du fait qu'une page est visitée avant une autre (l'ordre est souvent
une conséquence de la structure du site plutôt qu'un choix délibéré de
l'utilisateur).

D'autre part, nous simplifions les navigations en appliquant une
solution proposée dans \citet{Fu00} qui consiste à utiliser la
structure hiérarchique du site étudié. Une URL est en effet
organisée de façon hiérarchique : dans l'URL
\url{http://www-sop.inria.fr/axis/Publications/} choisie sur le
site de l'INRIA, on retrouve le serveur de l'unité de recherche de
l'INRIA située à Sophia-Antipolis (\url{www-sop.inria.fr}), le
projet de recherche AxIS (\url{axis}) et la liste de publications
de ses membres (\url{Publications}).  Pour simplifier l'analyse
d'un ensemble de navigations, on peut donc remplacer les URLs des
documents visités par une version ``raccourcie'' qui se base sur
la structure du site.

\begin{table}[htbp]
\begin{center}
  \begin{tabular}{|l|}\hline
 URL \\\hline\hline
 \url{http://www-sop.inria.fr/}\\\hline
 \url{http://www-sop.inria.fr/act_recherche/les_projets_fr.shtml}\\\hline
 \url{http://www.inria.fr/recherche/equipes/axis}\\\hline
 \url{http://www-sop.inria.fr/axis/}\\\hline
 \url{http://www-sop.inria.fr/axis/ra.html}\\\hline
 \url{http://www.inria.fr/rapportsactivite/RA2003/axis2003/axis_tf.html}\\\hline
  \end{tabular}
\end{center}
\centering \caption{Une navigation}\label{tabNavigation}
\end{table}

Considérons par exemple la navigation de la table
\ref{tabNavigation} (nous avons ici extrait la colonne URL d'un
tableau de données de la forme du tableau \ref{tableData}). Une
simplification possible de cette navigation consiste à ne
conserver que le nom du serveur et deux niveaux hiérarchique pour
chaque URL, ce qui donne la table \ref{tabNaviSimple}. En fait, on
remplace la variable URL du tableau de données d'origine par
plusieurs variables, une pour le serveur, puis une par niveau
hiérarchique conservé.

\begin{table}[htbp]
\begin{center}
  \begin{tabular}{|l|l|l|}\hline
Serveur & Niveau 1 & Niveau 2 \\\hline\hline
\url{www-sop.inria.fr}& & \\\hline
 \url{www-sop.inria.fr}&\url{act_recherche}&\url{les_projets_fr.shtml}\\\hline
 \url{www.inria.fr}&\url{recherche}&\url{equipes}\\\hline
 \url{www-sop.inria.fr}&\url{axis}&\\\hline
 \url{www-sop.inria.fr}&\url{axis}&\url{ra.html}\\\hline
 \url{www.inria.fr}&\url{rapportsactivite}&\url{RA2003}\\\hline
  \end{tabular}
\end{center}
\centering \caption{Une représentation simplifiée de la navigation de la table
\ref{tabNavigation}}\label{tabNaviSimple}
\end{table}

Notons que d'autres méthodes de regroupement d'URLs sont
envisageables, en travaillant par exemple sur le contenu des pages
ou encore sur la structure d'hyperlien du site. Cependant, il est
important de conserver des groupes facilement interprétables.
C'est le cas de notre méthode car le groupe d'URLs est obtenue
 par un simple élagage de l'arbre associé à l'URL.

\subsection{Une analyse de l'usage du site de l'INRIA}
\subsubsection{Les données}
Dans cette section, nous analysons l'usage d'une partie du site Web de
l'Institut National de Recherche en Informatique et Automatique (INRIA). Le
site de l'INRIA est reparti en plusieurs serveurs dont les rôles sont
différents. Le site principal, \url{www.inria.fr} présente l'institut dans son
ensemble, assure la diffusion des rapports de recherche, la promotion de
l'institut, etc. Les Unités de Recherche (UR) qui correspondent grossièrement
aux différentes implantations géographiques de l'INRIA possèdent aussi des
serveurs (il y a six unités de recherche). Nous nous sommes intéressés au
serveur de l'UR de Sophia Antipolis, \url{www-sop.inria.fr}. Comme l'illustre
la navigation de la table \ref{tabNavigation}, les différents serveurs de
l'INRIA sont étroitement liés et le passage de l'un d'entres eux à un autre se
fait de façon totalement transparente pour l'utilisateur. Une analyse
multi-serveurs est donc indispensable dans ce contexte.

Nous étudions les accès effectués sur les serveurs pendant les 15 premiers
jours de l'année 2003. Nous ne retenons que les longues navigations, c'est à
dire les navigations dont la durée est supérieur à $60$ secondes et dont le
nombre de pages visitées est supérieur à $10$ pages. De plus, nous ne nous
intéressons qu'aux navigations qui contiennent au moins une requête vers
chacun des deux sites. Au total, nous avons donc $3969$ navigations qui
correspondent à $282 552$ requêtes valides (statut entre $200$ et $399$). Dans
les analyses, nous appliquons les simplifications décrites dans la section
\ref{subsectionTopic} en ne conservant que le serveur et le premier niveau de
l'URL, que nous désignerons sous le terme de rubrique de niveau 1. Nous
obtenons ainsi 196 groupes d'URLs.

Notre analyse basée sur l'adaptation des cartes auto-organisatrices de Kohonen
aux tableaux de dissimilarités porte sur deux problèmes distincts à savoir :
\begin{itemize}
    \item l'analyse et la classification des navigations pour
    trouver des comportements types d'utilisateurs;
  \item l'analyse et la classification des rubriques de niveau 1 pour
    comprendre et analyser la perception du site par les internautes.
\end{itemize}

\subsubsection{Analyse des navigations}
Après la prise en compte de la structure du site, nous obtenons le tableau de
données de la table \ref{tableServeur}, dans laquelle ``www'' désigne le site
principal \url{www.inria.fr} et ``SOP'' le site de l'UR de Sophia Antipolis,
\url{www-sop.inria.fr}.

\begin{table}[htpb]
\begin{center}
\begin{tabular}{|c|c|c|c|}\hline
Requête&Navigation& Serveur& Rubrique 1  \\ \hline \hline
0& 1&www& robotvis \\
1& 1& SOP& robotvis \\
2&1& www& JGI2002 \\
$\vdots$&$\vdots$&$\vdots$&$\vdots$ \\
282 550& 3969& www& freesoft\\
282 551& 3969& SOP& freesoft\\ \hline
\end{tabular}
\end{center}
\caption{Les logs après simplification}\label{tableServeur}
\end{table}

Les observations de notre analyse sont les navigations : il nous
faut donc produire un nouveau tableau de données, où chaque ligne
contient la description d'une navigation. Comme nous ne tenons pas
compte du temps, nous décrivons chaque navigation par deux
variables modales. Chaque variable, les rubriques 1 de
\url{www.inria.fr} et de \url{www-sop.inria.fr}, est maintenant
représentée par une distribution de fréquences, comme le montre la
table \ref{tableModale}.

\begin{table}[htbp]
  \centering
  \begin{tabular}{|c|c|c|}\hline
     Navigation& Rubrique 1 sur www& Rubrique 1 sur SOP
    \\\hline
    1 &  robovis(10); JGI02(0),... &  robotvis(15), thesard(36),... \\
     \vdots & \vdots& \vdots \\

  3969 & rapport(63), axis(98),... & interne(64), saga(18), ... \\ \hline
  \end{tabular}
  \caption{Les navigations sous forme de variables modales}
  \label{tableModale}
\end{table}

Pour comparer deux navigations, nous utilisons le coefficient d'affinité
\citep{Mat51,Mat55,Bac85,Bac00}, dont nous rappelons la définition. Pour
chaque variable $Y_j$ à $t_j$ modalités, on suppose données les deux
distributions de fréquences, $\delta_{N_i^j}=(n_{ij1},..., n_{ij{t_j}})$ et
$\delta_{N_k^j}=(n_{kj1},..., n_{kj{t_j}})$, associées aux navigations $N_i$
et $N_k$. Le coefficient d'affinité est alors donné par :
\begin{equation}
  \label{eqAffinite}
  \aff(\delta_{N_i^j},\delta_{N_k^j})=\sum_{l=1}^{t_j}\sqrt{\frac{n_{ijl}}{n_{ij.}} \frac{n_{kjl}}{n_{kj.}}},
\end{equation}
avec $\displaystyle n_{ij.}=\sum_{l=1}^{t_j}n_{ijl}$ et  $n_{ijl}$
est le nombre d'occurrences pour la navigation $i$ dans la
modalité $l$ et la variable $Y_j$ ($1\leqslant i\leqslant 3969$ et
$1\leqslant l\leqslant t_j$).

Notons que $0\leqslant
\aff(\delta_{N_i^j},\delta_{N_k^j})\leqslant 1$. La valeur $1$ est
atteinte si $\delta_{N_i^j}$ et $\delta_{N_k^j}$ sont identiques
ou proportionnelles, alors que la valeur $0$ correspond au cas
d'orthogonalité.

Considérons maintenant $p$ variables modales (dans notre cas, $p=2$ et
correspond au nombre de serveurs). Soit $w_j$ le poids d'une variable $Y_j$
mesurant son importance, et tel que $0\leqslant w_j\leqslant 1$ et
$\displaystyle\sum_{j=1}^p w_j =1$. Nous définissons la similarité d'affinité
pondérée $a(N_i, N_k)$ entre deux navigations $N_i$ et $N_k$ par la moyenne
pondérée suivante :
\begin{equation}
  \label{eqAffPond}
 a(N_i, N_k)= \sum_{j=1}^p w_j
\aff(\delta_{N_i^j},\delta_{N_k^j})=\sum_{j=1}^p w_j
\sum_{l=1}^{t_j}\sqrt{\frac{n_{ijl}}{n_{ij.}}
\frac{n_{kjl}}{n_{kj.}}}
\end{equation}
La dissimilarité associée $d(N_i, N_k)$ entre deux navigations
$N_i$ et $N_k$ est alors définie comme suit :
\begin{equation}\label{eqDissNavi}
d(N_i, N_k)=2(1-a(N_i, N_k))
\end{equation}
Notre algorithme prend en entrée la matrice de dissimilarités basée sur le
coefficient d'affinité entre les $3969$ navigations.  La structure \emph{a
  priori} est celle d'une grille à deux dimensions, de taille $5 \times 4=20$
neurones. Chaque neurone est initialisé par un élément de l'ensemble
d'apprentissage $\Omega$ choisi aléatoirement (on a donc $q=1$). Le noyau $K$
est une Gaussienne (cf la section \ref{SOMClassique}). La
figure~\ref{carteNAVIG} représente la carte finale obtenue (le prefixe SOP-
signifie que la page a été consultée à partir du site de Sophia).

\begin{figure}[pbth]
\begin{center}
\begin{tiny}
\centering
\begin{tabular}{|p{2.2cm}|p{2.2cm}|p{2.2cm}|p{2.2cm}|p{2.2cm}|}
 \hline & & & &\\
\textbf{Recherche, inria, SOP-act-recherche, Travailler,
SOP-axis}&

\textbf{Recherche, Travailler, SOP-act-recherche, SOP-axis,
SOP-semir, actu}&

\textbf{Recherche, Travailler, SOP-act-recherche, SOP-semir,
SOP-actu, SOP-DR, interne, SOP-interne}, services, DR,
Relation-ext& \textbf{interne, SOP-interne, SOP-DR, SOP-semir}&
\textbf{interne, SOP-DR, SOP-interne, SOP-semir}, semir, DR\\ & & & & \\
 \emph{\normalsize{Classe 16}}& \emph{\normalsize{Classe 17}}&\emph{\normalsize{Classe 18}}& \emph{\normalsize{Classe 19}}&
\emph{\normalsize{Classe 20}}\\ \hline & & & &\\

  \textbf{Recherche, Valorisation, inria, rrrt, actualite, rapportactivite, cgi-bin, act-recherche, Travailler} personnel, sinus, 
SOP-sinus, SOP-miaou, SOP-omega, SOP-smash, SOP-caiman&

\textbf{Recherche, Travailler, inria,} SOP-lemme, SOP-Oasis&

\textbf{\textbf{DR}, Recherche, Travailler, inria, interne},
personnel, cermics, SOP-cermics, SOP-caiman& agos-sophia, acacia,
SOP-axis& \textbf{publication, dias, SOP-cgi-bin, SOP-dias,
SOP-interne}, SOP-actu
\\ & & & & \\
\emph{\normalsize{Classe 11}}& \emph{\normalsize{Classe 12}}&\emph{\normalsize{Classe 13}}& \emph{\normalsize{Classe 14}}& 
\emph{\normalsize{Classe 15}}\\ \hline & & & &\\

\textbf{Recherche, Valorisation, rrrt, rapportactivite,}
SOP-epidaure&

\textbf{Recherche, rapportactivite, rrrt, inria}, Robotvis,
SOP-Robotvis, SOP-Odyssee&

\textbf{rapportactivite, rrrt,} Prisme, SOP-Prisme&

\textbf{rrrt, Publications, SOP-cgi-bin, SOP-dias} &
\textbf{Publication, SOP-cgi-bin, dias, SOP-dias}
\\ & & & & \\
 \emph{\normalsize{Classe 6}}& \emph{\normalsize{Classe 7}}&\emph{\normalsize{Classe 8}}& \emph{\normalsize{Classe 9}}& 
\emph{\normalsize{Classe 10}}\\\hline & & & &\\

\textbf{Recherche, Valorisation, rrrt, rapportactivite,} Travailler, presse, personnel, inria, publications, actualite, multimedia, 
fonctions, SOP-robotvis, SOP-lemme, SOP-mistral&

\textbf{Recherche, rapportactivite,} icare, SOP-icare, RA95&

caiman, SOP-caiman, SOP-glaad, SOP-Safir, SOP-cgi-bin&

chir, SOP-chir, SOP-Saga& rrrt, icons, SOP-coprin \\ & & & & \\
\emph{\normalsize{Classe 1}}& \emph{\normalsize{Classe 2}}&\emph{\normalsize{Classe 3}}& \emph{\normalsize{Classe 4}}& 
\emph{\normalsize{Classe 5}} \\\hline
\end{tabular}
\end{tiny}
\caption{La carte finale $5 \times 4$ des navigations. Chaque
neurone contient le prototype associé (une navigation référente) et
donc la liste des rubriques 1 visitées par cette navigation dans les sites du
siège et de Sophia} \label{carteNAVIG}
\end{center}
\end{figure}

Chaque case de la carte contient les rubriques de niveau 1 visitées par la
navigation du prototype final. Les rubriques indiquées en gras mettent en
avant les ressemblances locales. Si nous prenons la classe 1 par exemple, les
rubriques 1 en gras sont : ``Recherche'', ``Valorisation'', ``rrrt'' et
``rapportactivite''. Les voisins directs de la classe 1 sont la classe 2 et la
classe 6. Les rubriques ``Recherche'' et ``rapportactivite'' ont été visitées
par ces trois classes. Les rubriques ``rrrt'' et ``Valorisation'' sont
communes à la classe 1 et la classe 6. La carte finale obtenue est
satisfaisante car les classes voisines partagent quelques rubriques. Il n'y a
donc pas lieu d'utiliser pour le paramètre $q$ (le nombre de prototype pour
chaque neurone) une valeur strictement supérieure à 1.

Cette première analyse des navigations indique des comportements types des
utilisateurs :
\begin{itemize}
\item le coin supérieur droit de la carte est consacré aux navigations sur des
  pages internes de l'INRIA (classes 19 et 20, avec les rubriques 1
  ``interne'',''SOP-interne'', ``semir'' un groupe de pages qui décrit les
  services informatiques internes, etc.) ;
\item la classe 18 réalise une transition entre la partie interne et des
  navigations consacrées à la recherche d'emploi à l'INRIA, c'est-à-dire les
  classes 11, 12, 13, 16 et 17 (Rubriques 1 ``Travailler'' et ``Recherche'') ;
\item le coin inférieur gauche de la carte est plutôt consacré à la
  présentation de la recherche (classes 1, 2, 6 et 7) par l'intermédiaire des
  rapports d'activités des équipes (rubriques 1 ``rapportactivite'' et
  ``rrrt''), ainsi que par la communication institutionnelle (rubriques 1
  ``actualite'', ``Valorisation'', ``Recherche'', ``presse'') ;
\item le coin inférieur droit de la carte est consacré à des navigations plus
  ciblées, portant sur des projets de recherche visités depuis leurs rapports
  techniques et d'activité.
\end{itemize}
Notons que nous nous sommes focalisés sur les navigations visitant
à la fois le site du siège de l'INRIA et celui de l'UR de Sophia.
On constate sur la carte que les liens entre les sites semblent
fonctionnels. Les navigations de recherche d'emploi (rubrique
``Travailler'') visitent souvent des sites de projet de recherche
de l'UR de Sophia, comme par exemple ``SOP-axis'' (classe 16),
``SOP-sinus'', ``SOP-Miaou'', etc. (classe 11) ou encore
``SOP-lemme'' et ``SOP-Oasis'' (classe 12). De façon générale,
excepté pour les classes correspondant à des navigations internes
(19 et 20), toutes les navigations référentes passent par des
sites de projets de recherche de l'UR de Sophia, ce qui valide le
rôle de promotion joué par le site du siège de l'INRIA.

\subsubsection{Analyse des rubriques}
La seconde analyse concerne les rubriques de niveau 1. Il s'agit
de déterminer comment les pages correspondantes sont perçues par
les utilisateurs du site. Pour cela, nous construisons un tableau
décrivant chaque navigation par la liste des rubriques de niveau 1
consultées. À partir de ce tableau, nous obtenons un tableau
binaire dont les individus sont les 196 rubriques de niveau 1 et
les variables sont les navigations : pour la rubrique $R_j$, la
variable $N_i$ vaut 1 si et seulement si la navigation $N_i$ a
visité au moins une page du groupe d'URLs décrit par la rubrique
$R_j$. Nous obtenons ainsi la table \ref{binaire}.

\begin{table}[htbp]
\centering
\begin{tabular}{|c c|c|c|c|c|}\hline
 & Navigations &$N_1$& $N_2$& ... &$N_{3969}$ \\
Rubriques &  & & &  &  \\ \hline \hline
$R_1=$ inria&  & 0 &1 &... & 0 \\
$R_2=$ Recherche&  & 1 &0 &... & 0 \\
$\vdots$& & $\vdots$& $\vdots$& $\vdots$& $\vdots$\\
$R_{196}=$ SOP-freesoft &  & 0 &0 &... & 0 \\ \hline
\end{tabular}
\caption{Tableau binaire décrivant les 196 rubriques en fonction des navigations} \label{binaire}
\end{table}

De nombreuses dissimilarités ont été définies pour les tableaux de données
binaires. Nous avons retenu celle basée sur l'indice de Jaccard car elle a
fait ses preuves dans le cadre de l'analyse de l'usage
\citep[cf][par exemple]{FossSDM2001}. Nous rappelons la définition de cette
dissimilarité.

Considérons deux vecteurs binaires $R_1$ et $R_2$ et introduisons les quatre
quantités suivantes :
\begin{itemize}
\item $a$ est le nombre de $j$ tels que $R_1^j= R_2^j=1$ ; \item
$b$ est le nombre de $j$ tels que  $R_1^j= 0$ et $ R_2^j=1$ ;
\item $c$ est le nombre de $j$ tels que  $R_1^j= 1$ et $ R_2^j=0$
; \item $d$ est le nombre de $j$ tels que $R_1^j= R_2^j=0$.
\end{itemize}
Ces définitions sont résumées par le tableau suivant :
\begin{center}
\begin{tabular}{|c c|c|c|}\hline
 & $R_1$ & $1$ &0 \\
$R_2$ &  & & \\ \hline \hline
1&  &$a$ & $b$ \\
0& & $c$ & $d$  \\ \hline
\end{tabular}
\end{center}
L'indice de Jaccard pour les vecteurs  $R_1$ et $R_2$ est donné par :
\begin{equation}
  \label{eqJaccard}
S(R_1, R_2)= \frac{a}{a+b+c}
\end{equation}
Dans notre contexte, il correspond à la probabilité de visite
de la rubrique $R_1$ et de la rubrique $R_2$ sachant qu'on a visité
au moins une des deux. La dissimilarité choisie est $(1-S)$, pour laquelle
nous appliquons donc l'algorithme proposé. Nous représentons ainsi 196
rubriques par une structure \emph{a priori} de grille de $4\times 3=12$
neurones.

Afin de faciliter l'analyse des résultats et de montrer leur pertinence, nous
avons enrichi la description des groupes d'URLs, c'est-à-dire les rubriques de
niveau 1, par une information sémantique obtenue par une analyse humaine du
site. Nous avons ainsi pu construire une taxonomie sur les
rubriques. Nous avons identifié des rubriques correspondant à des
manifestations (colloques, conférences, écoles d'été, etc.) et des rubriques
décrivant des projets de recherche. Les autres rubriques ont été classées en
rubriques inria ou sophia selon le serveur concerné.

De plus, les projets de recherche de l'INRIA étaient organisés en 2003 selon
quatre thèmes :
\begin{itemize}
\item thème 1 : réseaux et systèmes ;
\item thème 2 : génie logiciel et calcul symbolique ;
\item thème 3 : interaction homme-machine, images, données, connaissances ;
\item thème 4 : simulation et optimisation de systèmes complexes.
\end{itemize}
Ces thèmes permettent de subdiviser la catégorie projet.

\begin{figure}[htbp]
\centering\small
\begin{tabular}{|c|c|c|c|} \hline
manifestation& projet (thème 1)& projet (thème 3)& inria \\\hline
manifestation& projet (thème 1)& projet (thème 4)& projet (thème
2)\\\hline
projet (thème 2)& projet (thème 4) & projet (thème 4)& projet (thème 4) \\ \hline
\end{tabular}
\caption{La carte (4$ \times$ 3) obtenue : chaque case contient l'information
  sémantique associée à la rubrique référente}
\label{RubCarte}
\end{figure}

Nous présentons d'abord une vue de très haut niveau de la carte obtenue par
notre algorithme (voir la figure \ref{RubCarte}). Celle-ci est obtenue en
représentant l'information sémantique associée à la rubrique prototype de
chaque neurone. Nous constatons que l'organisation globale de la carte est
satisfaisante. En effet, les projets de thème 1 appartiennent à des classes
voisines, de même pour les projets de thème 4 ainsi que les manifestations. Il
apparaît donc que les utilisateurs du site sont soit convaincus par le
regroupement thématique des projets, soit contraints par la structure du site
à privilégier des visites communes à des projets dans les mêmes thèmes. En
fait, la réponse exacte est un mélange des deux interprétations. Dans les
navigations qui s'intéressent à plusieurs projets, il est fréquent de
retrouver des pages pivots qui présentent la recherche à l'INRIA. Ces pages
sont organisées par thème et induisent donc naturellement une navigation
thématique. Par contre, il est aussi fréquent de trouver des navigations qui
touchent plusieurs projets sans passage par des pages pivots. L'aspect
thématique est alors induit indirectement par des liens spécifiques (par
exemple des publications communes) ou par des ressources externes (par exemple
une recherche sur Google qui propose des pages provenant de différents
projets).

Nous représentons ensuite sur la figure \ref{CarteRubriqueProjet} le contenu
partiel de chaque classe. Plus précisément, nous indiquons l'affectation des
rubriques de niveau 1 classées dans la catégorie projet. La rubrique prototype
est indiquée en gras.

\begin{figure}[htbp]
\newcommand{\save}{\baselinestretch}
\renewcommand{\baselinestretch}{1}
\begin{scriptsize}
\begin{center}
\begin{sideways}
\begin{tabular}{|c|c|c|c|}
 \hline
  {\begin{tabular}{lp{2.2cm}}Thème 1& meije\\ Thème 2& Koala, croap \\ Thème 3& odyssee \\ Thème 4&Opale  \end{tabular}}&

{\begin{tabular}{lp{2.2cm}}\\Thème 1& SOP-mistral, SOP-Mimosa, SOP-sloop, SOP-rodeo, rodeo, mascotte, 
\textbf{SOP-mascotte}, sloop, SOP-planete, SOP-oasis
\end{tabular}}&

{\begin{tabular}{lp{2.2cm}}\\Thème 3& robotvis, epidaure, ariana, acacia, orion, aid, SOP-robotvis, SOP-epidaure, SOP-odyssee, 
SOP-acacia, \textbf{SOP-orion}, SOP-ariana, SOP-aid, SOP-axis, SOP-visa \end{tabular}}& \\
\textbf{Classe 9}& \textbf{Classe 10}&\textbf{Classe
11}&\textbf{Classe 12}\\ \hline

{\begin{tabular}{lp{2.2cm}}\\Thème 1& tropics\\ Thème 3&
reves
\\ Thème 4& Omega  \end{tabular}}&

{\begin{tabular}{lp{2.2cm}}\\Thème 1& Mimosa, tick,
\textbf{SOP-tick}\end{tabular}}&

{\begin{tabular}{lp{2.2cm}}\\Thème 4& \textbf{comore},
mefisto, miaou, SOP-mefisto, SOP-smash \end{tabular}}&

{\begin{tabular}{lp{2.2cm}}\\Thème 2& Prisme, SOP-Prisme, SOP-lemme, \textbf{SOP-galaad}, SOP-cafe, SOP-saga, 
SOP-safir\end{tabular}}\\
\textbf{Classe 5}& \textbf{Classe 6}&\textbf{Classe
7}&\textbf{Classe 8}\\ \hline

{\begin{tabular}{lp{2.2cm}}Thème 2& cafe, lemme,
\textbf{certilab}\\ Thème 4& Chir, Fractales, opale
\end{tabular}}&

{\begin{tabular}{lp{2.2cm}}\\Thème 1& Mistral, planete,  SOP-meije \\ Thème 2& oasis, saga, safir, SOP-Koala \\ Thème 4& 
\textbf{caiman}, sinus \end{tabular}}&

{\begin{tabular}{lp{2.2cm}}\\Thème 4& icare, SOP-sinus,
\textbf{SOP-icare}, SOP-miaou, SOP-caiman \end{tabular}}&

{\begin{tabular}{lp{2.2cm}}\\Thème 1& SOP-tropics\\ Thème 2& SOP-certilab\\ Thème 3& SOP-reves\\ Thème 4&\textbf{SOP-Omega}, 
SOP-sysdys \end{tabular}} \\
\textbf{Classe 1}& \textbf{Classe 2}&\textbf{Classe
3}&\textbf{Classe 4}\\ \hline
\end{tabular}
\end{sideways}
\end{center}
\end{scriptsize}
\centering
\caption{Affectation des projets aux différents neurones de la carte}
\label{CarteRubriqueProjet}
\renewcommand{\baselinestretch}{\save}
\end{figure}

Comme on peut le voir sur la carte détaillée, aucun projet n'a été affecté à
la classe 12, classe représentée par la rubrique sémantique "inria". Ceci
permet de déduire que la classe 12 est assez homogène. Il apparaît ainsi que
les pages des projets sont plutôt visitées indépendamment des autres pages des
serveurs.

Nous constatons aussi que les classes 3, 6, 7, 8, 10 et 11 sont composées
uniquement de projets appartenant aux même thèmes ce qui permet de déduire que
l'adaptation du SOM effectue une bonne quantification de l'espace. Il apparaît
aussi que les internautes semblent privilégier des navigations thématiques,
comme nous l'avons déjà remarqué pour la vue d'ensemble de la figure
\ref{RubCarte}.  Nous pouvons en outre constater pour la classe 11, constituée
de projets appartenant au thème 3, la présence simultanée du projet \emph{Aid}
et du projet \emph{Axis}.  Il faut savoir que le projet \emph{Axis} a remplacé
le projet \emph{Aid} au sein de l'INRIA. La visite de l'un entraîne donc très
souvent la visite de l'autre, car un lien mutuel existe entre les deux pages.
Nous retrouvons le même comportement pour le projet \emph{Odyssee} et le
projet \emph{Robotvis}.

Nous constatons la présence de projets dans les classes 5 et 9. En effet, ces
deux classes sont représentées par des manifestations et donc la présence des
projets permet de déduire que les manifestations sont liées à ces projets. Ce
qui explique la visite des pages des projets.

Un dernier point intéressant, si nous revenons à la première carte (figure
\ref{RubCarte}), nous constatons que certains projets de thème 2 appartiennent
à des classes très éloignées sur la carte.  Pour expliquer ce phénomène, il
faut savoir que:
\begin{itemize}
\item tout projet à l'INRIA a son propre site Web localisé sur le
serveur local de l'unité de recherche à laquelle il est rattaché ;
\item de plus, tout projet à l'INRIA possède une page descriptive
sur le serveur national du siège.
\end{itemize}
Prenons l'exemple du projet \emph{cafe}, qui est un projet de l'unité de
recherche de Sophia-Antipolis. Son site Web est donc localisé sur le serveur
de l'unité de recherche de Sophia, que nous avons noté \emph{SOP-cafe} sur la
carte. La page descriptive du projet \emph{cafe} est quant à elle localisée
sur le serveur national du siège (elle est désignée par \emph{cafe} sur la
carte). Nous nous attendons à ce que ces deux rubriques, \emph{cafe} et
\emph{SOP-cafe}, apparaissent dans la même classe ou dans des classes
voisines, car elles sont liées sémantiquement. Or, comme on peut le constater
sur la carte, la rubrique \emph{cafe} appartient à la classe 1 et la rubrique
\emph{SOP-cafe} appartient à la classe 8 qui ne sont pas voisines. Ce
phénomène peut être expliqué par l'absence de lien dans la page descriptive
vers le site Web du projet \emph{cafe}.  Donc au cours d'une même navigation,
l'internaute peut difficilement passer de la page descriptive vers le site Web
du projet, ce qui démontre un défaut de conception du site. Nous remarquons le
même comportement pour le projet \emph{lemme}.

\section{Conclusion}
Dans ce travail nous avons proposé une adaptation des cartes
auto-organisatrices de Kohonen aux tableaux de dissimilarités. Cette
adaptation est basée sur la version \emph{batch} de l'algorithme et permet de
traiter tout type de données. Les expériences ont montré l'efficacité de cette
méthode et son adaptation aux divers données complexes dès lors que l'on peut
définir une mesure de dissimilarité. Cette méthode a aussi donné de bons
résultats sur d'autres applications réelles et pour d'autres types de données
complexes \citep[voir][]{these04,ElGolliConanGuezRossi04JSDA}.

\section*{Remerciements}
Nous remercions Brigitte Trousse, Doru Tanasa et Mihai Jurca (équipe AxIS INRIA
Sophia-Antipolis) pour leur travail de pré-traitement sur les données analysées
dans cet article. Nous remercions aussi le rapporteur anonyme dont les
remarques et conseils ont contribué à améliorer le présent article.

\end{document}